\documentclass[
]{ceurart}

\usepackage{listings}
\begin{document}

\copyrightyear{2025}
\copyrightclause{Copyright for this paper by its authors.
  Use permitted under Creative Commons License Attribution 4.0
  International (CC BY 4.0).}

\conference{}

\title{Measuring Gender Bias in Job Title Matching for Grammatical Gender Languages}





\author{Laura Garc{\'\i}a-Sardi{\~n}a}[%
orcid={0000-0003-4592-8884},
email={machinelearning@avature.net}
]

\author{Hermenegildo Fabregat}[
orcid={0000-0001-9820-2150}
]

\author{Daniel Deniz}[
orcid=0000-0002-0313-2127,
]

\author{Rabih Zbib}[
orcid=0000-0002-7140-3048,
]

\address{Avature Machine Learning, Spain
}




\begin{abstract}
This work sets the ground for studying how explicit grammatical gender assignment in job titles can affect the results of automatic job ranking systems.
We propose the usage of metrics for ranking comparison controlling for gender to evaluate gender bias in job title ranking systems, in particular RBO (Rank-Biased Overlap). We generate and share test sets for a job title matching task in four grammatical gender languages, including occupations in masculine and feminine form and annotated by gender and matching relevance. We use the new test sets and the proposed methodology to evaluate the gender bias of several out-of-the-box multilingual models to set as baselines, showing that all of them exhibit varying degrees of gender bias.
\end{abstract}


\maketitle

\section{Introduction}

To address the challenges of today's fast-paced job market, organisations increasingly rely on automatic job recommendation systems to match job openings and candidates based on information like job titles, education, skills, or combinations of these \cite{decorte2021jobbert,zhao2021embedding,zbib2022learning,deniz2024}. 
These systems typically use Natural Language Processing (NLP) models to produce their recommendations.
However, as studies have shown that NLP systems can perpetuate and even amplify existing sociodemographic biases, such as gender bias, there is growing concern about their impact on real-world applications. 

\textit{Fairness} has been given several definitions in the literature. In this work, we adhere to the definition of \textit{fairness through awareness}~\cite{grgic2016case,chen2024fairness}, which requires systems to produce similar outcomes for similar individuals that differ only in some \textit{sensitive} or \textit{protected attribute(s)}, in this case gender. In job title matching systems in particular, a \textit{fair} system would give the same opportunities to people with the same occupation independently of their gender. 

Natural languages vary with respect to how they encode gender. Depending on their degree of grammatical gender marking, \citet{stahlberg2007} propose a categorisation of languages as: (a) \textit{grammatical gender} languages, like Spanish or German, in which every noun has a gender value, and other parts of speech can also be gender-marked morphologically (e.g. `peluquera'/`peluquero', `Friseurin'/`Friseur'); (b) \textit{natural gender} languages, like English or Swedish, in which gender is made explicit mostly in personal pronouns (e.g. `she'/`he') or lexically (e.g. `queen'/`king'); and (c) \textit{genderless} languages, like Turkish or Finnish, for which gender is not marked for nouns nor pronouns and can only be expressed by lexical means.

This work focuses on gender-marking languages and addresses how explicit gender assignment in job titles can affect the results of automatic job ranking systems. In this setting, \textit{gender bias} is understood as obtaining different recommended job rankings for the same query job title when the only difference is its morphological gender mark, e.g. `profesor' and `profesora' (masc. and fem. `teacher') obtaining different job recommendations. This scenario has an impact on the real world as an \textit{allocational harm}~\cite{blodgett2020language}, as such a system would allocate different opportunities (i.e. jobs) to candidates based on the gender expressed by their job title. To evaluate gender bias in job title ranking systems, we propose the use of metrics to compare rankings, controlling for query gender.

The main contributions of our work are:
\begin{itemize}
    \item The publication of new test sets for job title similarity with gender annotations, in four gender-marking languages.
    \item The proposal of a methodology to measure gender bias in job title ranking systems for grammatical gender languages, using Rank-Biased Overlap (RBO) to compare rankings controlling for gender.
    \item The evaluation of gender bias of some publicly available out-of-the-box models over the new test sets.
\end{itemize}

Our paper seeks to encourage further research in this field by introducing new benchmarks and offering some preliminary baseline explorations.

\section{Related Work}

In recent years, gender bias has become a well-documented and researched topic in NLP. Studies have proven that pre-trained language models show biases in their representations, through the Word Embedding Association Test (WEAT)~\cite{caliskan2017semantics,caliskan2022gender} and the later Sentence Encoder Association Test (SEAT)~\cite{may2019measuring}, or tasks like co-reference resolution \cite{zhao2018gender,rudinger2018gender} and natural language inference \cite{sharma2021evaluating}. 
Occupations have also been used in studying gender bias in NLP models and applications, particularly for stereotypes \cite{bhaskaran2019good,bolukbasi2016man,rudinger2018gender}. 
However, most of the research has focused only on English. When expanding their work to German, a language with grammatical gender, \citet{bartl2020unmasking} showed that techniques previously used on English to measure and mitigate gender bias were not applicable to German. Although gender bias might be even more notable for languages with grammatical gender, research has been scarce, and often limited to Machine Translation (MT) from English to gender marking languages \cite{piazzolla2023good}, or in cross-lingual embedding comparison settings \cite{zhao2020gender}. 
More recently, \citet{zhao2024gender} studied gender bias in Large Language Models (LLMs) across multiple languages, including grammatical gender ones. By analysing co-occurrence between certain words and genders, and the prediction of gender roles (e.g. through pronouns) from a description, and sentiment from generated dialogues, they detected gender bias in all the tested languages.

In this study, we build upon earlier research by introducing a novel approach to assessing gender bias in ranking situations involving languages with grammatical gender. We apply this method to a downstream task with significant real-world implications --specifically, job title matching-- 
and establish preliminary baselines using models that are currently feasible and comply with practical constraints for industrial application.

\section{Methodology}

\begin{figure}
  \centering
  \includegraphics[width=0.7\linewidth]{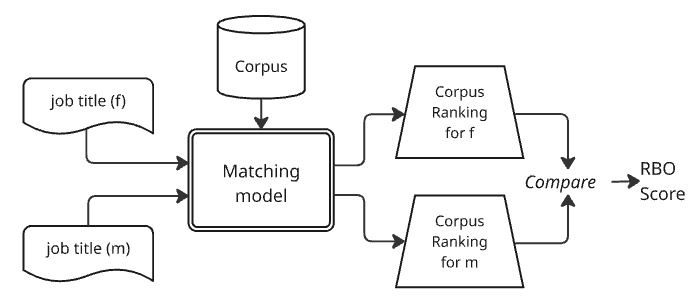}
  \caption{Overview of the proposed methodology for measuring gender bias in job title matching systems. The input is the same job title in feminine and in masculine forms, which are matched against the same corpus and then have their rankings compared. The process is repeated over a set of test query job titles, having their RBO scores averaged to get the final system's bias metric.}
  \label{fig:methodology}
\end{figure}

We address the job recommendation problem as an Information Retrieval task where, given some input job title or \textit{query}, we rank by relevance the items in a \textit{corpus} or pool of candidate job titles. Then, to assess gender bias, we measure the similarity between the generated ranks when we use feminine versus masculine queries over the same corpus. An overview of the methodology is depicted in Figure~\ref{fig:methodology}.

\subsection{Datasets}

In this work, we compile new test sets of job titles annotated for job title ranking and including marked gender information for each occupation. For this purpose, we use the English job title similarity dataset from \citet{zbib2022learning}, which is already annotated for the job matching task and is publicly available\footnote{\url{https://github.com/rabihzbib/jobtitlesimilarity_dataset}}. We automatically translate the occupations in this dataset into a set of Indo-European languages with grammatical gender: Spanish, German, French, and Portuguese. 

The phenomenon of gender bias in MT systems has been widely analysed (\citealp{stanovsky2019evaluating,prates2020assessing,savoldi2021gender,piazzolla2023good}, among others). Given a list of occupations as input to translate from a non-gender-marked language to a gender-marked one, MT systems will output a translation for each of them choosing one gender, typically replicating human stereotypes. 
As \citet{sharma2021evaluating} point out, adding an unambiguous context can \textit{fix} the MT system's gender bias. Consequently, with the goal of generating alternative masculine and feminine translations of occupations, we use contextual templates to induce translation into a target gender, similarly to \citet{cho2019measuring}: \texttt{`He is: \{job\_title\}.'} and \texttt{`She is: \{job\_title\}.'}\footnote{Several templates were tried over a few samples and evaluated qualitatively before finally selecting these.}. The resulting translation is post-processed to remove the template text and leave only the job title in the target language, keeping track of the assigned gender.

Other post-processing steps we perform include converting to lowercase and removing repeated terms (i.e., we only keep one when two or more source terms produce the same translation in the target language, 
e.g. feminine `lawyer' and `attorney' are both translated as `abogada' in Spanish). We also merge translations where the feminine and masculine terms are equal, tagging them as \textit{neutral}, as these are job titles without explicitly marked gender and one would have to rely on additional information or fall into stereotypes to assume the gender of the person in that position. For example, the translation into Portuguese of `data analyst' would result in `analista de dados' in both its masculine and feminine forms, so we consider it a neutral term.  

The automatic translations produced for the Spanish test set were reviewed and manually corrected. For the rest of the languages, the translations produced by Google Translate\footnote{\url{https://translate.google.com/}} were kept as such.  
Table~\ref{tab:stats} presents some statistics on the final test sets, which we make publicly available\footnote{Available upon request.} and includes annotations at both the gender and matching relevance levels.

\begin{table}
\centering
\caption{Number of masculine (M) and feminine (F) pairs, neutral (N), and total (T) job titles per language and set. The total differs across languages because of repetition removal and merged neutral 
terms.
}
\label{tab:stats}
\begin{tabular}{|l|ccc|ccc|}
\hline
\multirow{2}{*}{\textbf{Set}} & \multicolumn{3}{c|}{\textbf{Queries}}                                            & \multicolumn{3}{c|}{\textbf{Corpus}}                                             \\ \cline{2-7} 
                              & \multicolumn{1}{c|}{\textbf{M/F}} & \multicolumn{1}{c|}{\textbf{N}} & \textbf{T} & \multicolumn{1}{c|}{\textbf{M/F}} & \multicolumn{1}{c|}{\textbf{N}} & \textbf{T} \\ \hline
\textbf{DE}                   & \multicolumn{1}{c|}{99}           & \multicolumn{1}{c|}{5}          & 203        & \multicolumn{1}{c|}{2264}         & \multicolumn{1}{c|}{201}        & 4729       \\ \hline
\textbf{ES}                   & \multicolumn{1}{c|}{81}           & \multicolumn{1}{c|}{23}         & 185        & \multicolumn{1}{c|}{2052}         & \multicolumn{1}{c|}{557}        & 4661       \\ \hline
\textbf{FR}                   & \multicolumn{1}{c|}{60}           & \multicolumn{1}{c|}{44}         & 164        & \multicolumn{1}{c|}{1566}         & \multicolumn{1}{c|}{985}        & 4117       \\ \hline
\textbf{PT}                   & \multicolumn{1}{c|}{75}           & \multicolumn{1}{c|}{29}         & 179        & \multicolumn{1}{c|}{1703}          & \multicolumn{1}{c|}{899}       & 4305       \\ \hline
\end{tabular}
\end{table}

\subsection{Bias Metric}

We propose using Rank-Biased Overlap (RBO)~\cite{webber2010similarity} to assess the gender bias of job title rankers. RBO was originally developed to measure the similarity between ranked lists by assigning exponentially decreasing weights to items as their rank increases. This metric is particularly well suited for our purpose because it emphasises the agreement at the top of ranked lists while still considering the full ordering. RBO is normalised to a range from 0 to 1, where a value of 1 means complete concordance between the two lists and 0 indicates total dissimilarity. Equation~\ref{eq:rbo} shows the RBO formula:

\begin{equation}
\label{eq:rbo}
RBO = \frac{1}{k}\sum_{d=1}^{k} \frac{|S_{:d} \cap T_{:d}|}{d}    
\end{equation}

\noindent where $S$ and $T$ are the two rankings to be compared 
and $k$ is the evaluation depth. 
We consider the full rank, so the depth is equal to the length of the lists. 

In our study, we generate two parallel ranking scenarios. In one scenario, all the queries are in feminine form, while in the other they are in masculine form. The retrieval corpus remains constant between both scenarios to ensure that any differences in the rankings can be solely attributed to the gendered nature of the queries. For each job title, we compute the RBO between the corpus rankings that correspond to the feminine and masculine queries. We then average these scores across all queries by gender. An average RBO score closer to 1 suggests that the gender marking in the queries has little effect on the ranking outcome, whereas a lower score indicates a significant impact and, consequently, a stronger presence of gender bias. Note that in our results the pairs of rankings being compared come from the same corpora, meaning that items in both lists are equal. In this case, the lowest RBO score, which is over a list and its reverse is 0.5.  

\subsection{Selected Models}

In order to validate our proposed approach and to set a starting point for future research, 
we select different multilingual models and assess the impact on the resulting rankings based on the gender of the query. 

The selected models are the following: Multilingual Universal Sentence Encoder~\cite{yang2020multilingual} based on CNN (m-use) and Transformers (m-use-l), Multilingual E5~\cite{wang2024multilingual} base (m-e5) and large (m-e5-l), and Paraphrase Multilingual MiniLM-L12~\cite{reimers-2019-sentence-bert} (minilm). While m-use is a CNN-based architecture, the rest of the selected models are Transformer-based. All these models are pre-trained for multilingual semantic text similarity. We evaluate them out-of-the-box without any fine-tuning.

We selected these models based on their multilingual capabilities, their semantic similarity pre-training objective, and their size, as these characteristics make them eligible for their deployment in realistic application scenarios.

\section{Results and Analysis}
\label{sec:results}

Table~\ref{tab:results} shows the RBO results of the above models for the job title ranking task, comparing when we use feminine versus masculine queries over the same corpora. In this case, the reported corpora includes masculine forms, 
results over feminine corpora can be found in Table~\ref{tab:appendix_results_rbo} (Appendix~\ref{sec:appendix}).

The tested models present varying levels of gender bias, 
with differences up to four points across the evaluated languages depending on the model. This means that some of these base models present stronger gender bias for particular languages. For example, minilm is more sensitive to gender marking for German than for the other languages evaluated.
Additionally, analysing the influence of the size and complexity of the models, we see that models such as m-e5 and m-use demonstrate stability and reliability, especially in their larger configurations, generating more balanced and consistent results.
It must be noted that, in this use case, the RBO scores are obtained over very long rankings. Having all the models and languages give scores several points below 1 indicates that the order on the high ranking elements is significantly affected by the gender difference in the input queries. 

We emphasise that the presented methodology intends to measure gender bias in the proposed scenario, but does not substitute other evaluations of the performance of the models and the quality of their ranking capabilities. To demonstrate this, we compute the Mean Average Precision (MAP) of the selected models over our four test sets. 
Averaging scores over the four languages, results show that, while minilm exhibits the smallest gender bias, it is the worst performing option of the models tested (0.3382 MAP versus 0.37261 for m-e5, 0.3901 for m-e5-l, 0.3664 for m-use, and 0.3591 for m-use-l, over the masculine corpora).
Detailed MAP results per language and query type are reported in Tables~\ref{tab:map_m} and \ref{tab:map_f}, over masculine and feminine corpora respectively, in Appendix~\ref{sec:appendix}.

Finding a trade-off between model performance and gender bias is an important issue to address when developing and selecting job matching models for deployment. On the one hand, choosing a model with apparent good performance but that in turn shows a considerable gender gap may not only be ethically questionable, but it may also result in reputation and even legal consequences on the company responsible for it. On the other hand, selecting a model with minimal or no gender bias but with low quality may be considered to not fulfil its original aim. To develop fair and effective job matching models, it is imperative to address both aspects. Besides addressing them directly during training or in pre- or post-processing stages, e.g. using gender bias mitigation strategies, it becomes necessary to evaluate how models behave on both task performance and gender bias.


\begin{table}
\caption{RBO results of zero-shot performance of selected models, comparing feminine queries versus masculine queries over masculine corpora. Higher scores, in bold, mean less gender impact, and lower scores, underlined, a higher impact of changing the queries' gender.} 
\label{tab:results}
\begin{tabular}{|l|c|c|c|c|}
\hline
\textbf{Model} & \textbf{DE}     & \textbf{ES}     & \textbf{FR}     & \textbf{PT}              \\ \hline
m-e5           & \underline{0.8896} & \underline{0.9060} & \underline{0.9270} & 0.9236                   \\ \hline
m-e5-l         & 0.8966          & 0.9113          & 0.9391          & 0.9243                   \\ \hline
m-use          & 0.9128          & 0.9077          & \underline{0.9270} & \underline{0.9016}          \\ \hline
m-use-l        & \textbf{0.9254} & 0.9140          & 0.9435          & 0.9294                   \\ \hline
minilm         & 0.9095          & \textbf{0.9436} & \textbf{0.9448} & \textbf{0.9510} \\ \hline

\end{tabular}
\end{table}

In order to do a qualitative analysis of the results, we select one of the models and get the ranking for the feminine and masculine queries pair with the lowest RBO results for a given language, denoting the largest gender bias. 
We focus on Spanish, since that test set has been manually corrected, and 
choose the m-e5 model. %
The lowest RBO score corresponds to the query for `lawyer' (`abogada', `abogado'). When we inspect the top 20 ranked job titles for the feminine and masculine forms and compare them, the differences are notable. For the masculine query, only two job titles with a word meaning `associate' and `assistant' made it to the top 20, while the rest refer to lawyer specialisations (`abogado litigante', `abogado de empresa', `abogado penalista'...). For the feminine query, on the other hand, eight job titles out of 20 contained results with words like `adjunto' (`associate'), `ayudante'/`asistente' (`assistant'), and even `becario' (`intern'), therefore placing job positions of a lower seniority higher in the rank for the feminine input. Additionally, one of the job titles had no direct relation with the query. The lists with the top 20 ranked results for this query in feminine versus masculine are reported in Section~\ref{sec:appendix_b}.

\section{Conclusion}

In this work, we propose a methodology to measure gender bias in a high-impact NLP application in the human resources domain: job title matching. Using an existing test set in English for this task, we have generated gender-annotated analogous corpora in four languages with grammatical gender, and addressed the evaluation of gender bias as ranking comparison controlling for gender. 
Additionally, we establish baselines and confirm that this type of bias already exists in out-of-the-box pre-trained models, which are often used as the core for developing job title matching applications.
This stresses the need to identify and mitigate gender bias in order to create fair systems that give equal opportunities to users regardless of their gender. 
Future research may take advantage of the proposed test sets and evaluation methodology to assess gender bias mitigation techniques for grammatical gender languages.

\section*{Limitations}  

In our study, we focus on traditional feminine and masculine linguistic forms as explicitly expressed by morphological gender marks in job titles across several grammatical gender languages. Although this work addresses bimodal forms, we acknowledge and emphasise that gender is complex and encompasses a diverse range of biosocial factors distinct from biological sex. Natural languages, including grammatical gender ones, are evolving with new linguistic forms to recognise the broader non-binary nature of gender, sometimes based on new morphological markers (e.g. Portuguese and Spanish \textit{-e} in addition to \textit{-a} and \textit{-o}, `abogade') or on new orthographic rules (e.g. German introducing a colon before the gender-specific suffix, `Student:in'). One limitation of our work is that it focuses on morphologically marked bimodal gender forms, failing to cover other genders that do not fit into those representations. Our approach to generate gender-induced translations using current MT systems falls short to consider non-binary alternatives, as such systems would not produce such generations, even if we used a gender-neutral template (`They are: \{job\_title\}' would produce a plural translation with stereotyped gender).  

\section*{Declaration on Generative AI}
  

 During the preparation of this work, the authors used the suggestions provided by Writefull's model in its integration to Overleaf for grammar and spelling checking. Suggestions were always reviewed for acceptance or rejection, and the authors take full responsibility for the content of the publication.

\bibliography{paper}

\begin{thebibliography}{28}
\expandafter\ifx\csname natexlab\endcsname\relax\def\natexlab#1{#1}\fi
\providecommand{\url}[1]{\texttt{#1}}
\providecommand{\href}[2]{#2}
\providecommand{\path}[1]{#1}
\providecommand{\DOIprefix}{doi:}
\providecommand{\ArXivprefix}{arXiv:}
\providecommand{\URLprefix}{URL: }
\providecommand{\Pubmedprefix}{pmid:}
\providecommand{\doi}[1]{\href{http://dx.doi.org/#1}{\path{#1}}}
\providecommand{\Pubmed}[1]{\href{pmid:#1}{\path{#1}}}
\providecommand{\bibinfo}[2]{#2}
\ifx\xfnm\relax \def\xfnm[#1]{\unskip,\space#1}\fi
\bibitem[{Decorte et~al.(2021)Decorte, Van~Hautte, Demeester, and Develder}]{decorte2021jobbert}
\bibinfo{author}{J.-J. Decorte}, \bibinfo{author}{J.~Van~Hautte}, \bibinfo{author}{T.~Demeester}, \bibinfo{author}{C.~Develder},
\newblock \bibinfo{title}{Jobbert: understanding job titles through skills},
\newblock in: \bibinfo{booktitle}{FEAST2021, the International Workshop on Fair, Effective And Sustainable Talent management using data science}, \bibinfo{year}{2021}.
\bibitem[{Zhao et~al.(2021)Zhao, Wang, Sigdel, Zhang, Hoang, Liu, and Korayem}]{zhao2021embedding}
\bibinfo{author}{J.~Zhao}, \bibinfo{author}{J.~Wang}, \bibinfo{author}{M.~Sigdel}, \bibinfo{author}{B.~Zhang}, \bibinfo{author}{P.~Hoang}, \bibinfo{author}{M.~Liu}, \bibinfo{author}{M.~Korayem},
\newblock \bibinfo{title}{Embedding-based recommender system for job to candidate matching on scale},
\newblock \bibinfo{journal}{arXiv preprint arXiv:2107.00221}  (\bibinfo{year}{2021}).
\bibitem[{Zbib et~al.(2022)Zbib, Lacasa, Retyk, Poves, Aizpuru, Fabregat, Simkus, and Garc{\'\i}a-Casademont}]{zbib2022learning}
\bibinfo{author}{R.~Zbib}, \bibinfo{author}{L.~A. Lacasa}, \bibinfo{author}{F.~Retyk}, \bibinfo{author}{R.~Poves}, \bibinfo{author}{J.~Aizpuru}, \bibinfo{author}{H.~Fabregat}, \bibinfo{author}{V.~Simkus}, \bibinfo{author}{E.~Garc{\'\i}a-Casademont},
\newblock \bibinfo{title}{Learning job titles similarity from noisy skill labels},
\newblock \bibinfo{journal}{arXiv preprint arXiv:2207.00494}  (\bibinfo{year}{2022}).
\bibitem[{Deniz et~al.(2024)Deniz, Retyk, García-Sardiña, Fabregat, Gasco, and Zbib}]{deniz2024}
\bibinfo{author}{D.~Deniz}, \bibinfo{author}{F.~Retyk}, \bibinfo{author}{L.~García-Sardiña}, \bibinfo{author}{H.~Fabregat}, \bibinfo{author}{L.~Gasco}, \bibinfo{author}{R.~Zbib},
\newblock \bibinfo{title}{Combined unsupervised and contrastive learning for multilingual job recommendation},
\newblock in: \bibinfo{booktitle}{CEUR Workshop Proceedings}, volume \bibinfo{volume}{3788}, \bibinfo{publisher}{CEUR-WS}, \bibinfo{year}{2024}.
\bibitem[{Grgic-Hlaca et~al.(2016)Grgic-Hlaca, Zafar, Gummadi, and Weller}]{grgic2016case}
\bibinfo{author}{N.~Grgic-Hlaca}, \bibinfo{author}{M.~B. Zafar}, \bibinfo{author}{K.~P. Gummadi}, \bibinfo{author}{A.~Weller},
\newblock \bibinfo{title}{The case for process fairness in learning: Feature selection for fair decision making},
\newblock in: \bibinfo{booktitle}{NIPS Symposium on Machine Learning and the Law}, volume~\bibinfo{volume}{1}, \bibinfo{organization}{Barcelona, Spain}, \bibinfo{year}{2016}, p.~\bibinfo{pages}{11}.
\bibitem[{Chen et~al.(2024)Chen, Zhang, Hort, Harman, and Sarro}]{chen2024fairness}
\bibinfo{author}{Z.~Chen}, \bibinfo{author}{J.~M. Zhang}, \bibinfo{author}{M.~Hort}, \bibinfo{author}{M.~Harman}, \bibinfo{author}{F.~Sarro},
\newblock \bibinfo{title}{Fairness testing: A comprehensive survey and analysis of trends},
\newblock \bibinfo{journal}{ACM Transactions on Software Engineering and Methodology} \bibinfo{volume}{33} (\bibinfo{year}{2024}) \bibinfo{pages}{1--59}.
\bibitem[{Stahlberg et~al.(2007)Stahlberg, Braun, Irmen, and Sczesny}]{stahlberg2007}
\bibinfo{author}{D.~Stahlberg}, \bibinfo{author}{F.~Braun}, \bibinfo{author}{L.~Irmen}, \bibinfo{author}{S.~Sczesny},
\newblock \bibinfo{title}{Representation of the sexes in language},
\newblock \bibinfo{journal}{Social Communication}  (\bibinfo{year}{2007}) \bibinfo{pages}{163--187}.
\bibitem[{Blodgett et~al.(2020)Blodgett, Barocas, Daum{\'e}~III, and Wallach}]{blodgett2020language}
\bibinfo{author}{S.~L. Blodgett}, \bibinfo{author}{S.~Barocas}, \bibinfo{author}{H.~Daum{\'e}~III}, \bibinfo{author}{H.~Wallach},
\newblock \bibinfo{title}{Language (technology) is power: A critical survey of “bias” in nlp},
\newblock in: \bibinfo{booktitle}{Proceedings of the 58th Annual Meeting of the Association for Computational Linguistics}, \bibinfo{year}{2020}, pp. \bibinfo{pages}{5454--5476}.
\bibitem[{Caliskan et~al.(2017)Caliskan, Bryson, and Narayanan}]{caliskan2017semantics}
\bibinfo{author}{A.~Caliskan}, \bibinfo{author}{J.~J. Bryson}, \bibinfo{author}{A.~Narayanan},
\newblock \bibinfo{title}{Semantics derived automatically from language corpora contain human-like biases},
\newblock \bibinfo{journal}{Science} \bibinfo{volume}{356} (\bibinfo{year}{2017}) \bibinfo{pages}{183--186}.
\bibitem[{Caliskan et~al.(2022)Caliskan, Ajay, Charlesworth, Wolfe, and Banaji}]{caliskan2022gender}
\bibinfo{author}{A.~Caliskan}, \bibinfo{author}{P.~P. Ajay}, \bibinfo{author}{T.~Charlesworth}, \bibinfo{author}{R.~Wolfe}, \bibinfo{author}{M.~R. Banaji},
\newblock \bibinfo{title}{Gender bias in word embeddings: A comprehensive analysis of frequency, syntax, and semantics},
\newblock in: \bibinfo{booktitle}{Proceedings of the 2022 AAAI/ACM Conference on AI, Ethics, and Society}, \bibinfo{year}{2022}, pp. \bibinfo{pages}{156--170}.
\bibitem[{May et~al.(2019)May, Wang, Bordia, Bowman, and Rudinger}]{may2019measuring}
\bibinfo{author}{C.~May}, \bibinfo{author}{A.~Wang}, \bibinfo{author}{S.~Bordia}, \bibinfo{author}{S.~Bowman}, \bibinfo{author}{R.~Rudinger},
\newblock \bibinfo{title}{On measuring social biases in sentence encoders},
\newblock in: \bibinfo{booktitle}{Proceedings of the 2019 Conference of the North American Chapter of the Association for Computational Linguistics: Human Language Technologies, Volume 1 (Long and Short Papers)}, \bibinfo{year}{2019}, pp. \bibinfo{pages}{622--628}.
\bibitem[{Zhao et~al.(2018)Zhao, Wang, Yatskar, Ordonez, and Chang}]{zhao2018gender}
\bibinfo{author}{J.~Zhao}, \bibinfo{author}{T.~Wang}, \bibinfo{author}{M.~Yatskar}, \bibinfo{author}{V.~Ordonez}, \bibinfo{author}{K.-W. Chang},
\newblock \bibinfo{title}{Gender bias in coreference resolution: Evaluation and debiasing methods},
\newblock in: \bibinfo{booktitle}{Proceedings of the 2018 Conference of the North American Chapter of the Association for Computational Linguistics: Human Language Technologies}, volume~\bibinfo{volume}{2}, \bibinfo{year}{2018}.
\bibitem[{Rudinger et~al.(2018)Rudinger, Naradowsky, Leonard, and Van~Durme}]{rudinger2018gender}
\bibinfo{author}{R.~Rudinger}, \bibinfo{author}{J.~Naradowsky}, \bibinfo{author}{B.~Leonard}, \bibinfo{author}{B.~Van~Durme},
\newblock \bibinfo{title}{Gender bias in coreference resolution},
\newblock in: \bibinfo{booktitle}{Proceedings of the 2018 Conference of the North American Chapter of the Association for Computational Linguistics: Human Language Technologies, Volume 2 (Short Papers)}, \bibinfo{year}{2018}, pp. \bibinfo{pages}{8--14}.
\bibitem[{Sharma et~al.(2021)Sharma, Dey, and Sinha}]{sharma2021evaluating}
\bibinfo{author}{S.~Sharma}, \bibinfo{author}{M.~Dey}, \bibinfo{author}{K.~Sinha},
\newblock \bibinfo{title}{Evaluating gender bias in natural language inference},
\newblock \bibinfo{journal}{arXiv preprint arXiv:2105.05541}  (\bibinfo{year}{2021}).
\bibitem[{Bhaskaran and Bhallamudi(2019)}]{bhaskaran2019good}
\bibinfo{author}{J.~Bhaskaran}, \bibinfo{author}{I.~Bhallamudi},
\newblock \bibinfo{title}{Good secretaries, bad truck drivers? occupational gender stereotypes in sentiment analysis},
\newblock in: \bibinfo{booktitle}{Proceedings of the First Workshop on Gender Bias in Natural Language Processing}, \bibinfo{year}{2019}, pp. \bibinfo{pages}{62--68}.
\bibitem[{Bolukbasi et~al.(2016)Bolukbasi, Chang, Zou, Saligrama, and Kalai}]{bolukbasi2016man}
\bibinfo{author}{T.~Bolukbasi}, \bibinfo{author}{K.-W. Chang}, \bibinfo{author}{J.~Y. Zou}, \bibinfo{author}{V.~Saligrama}, \bibinfo{author}{A.~T. Kalai},
\newblock \bibinfo{title}{Man is to computer programmer as woman is to homemaker? debiasing word embeddings},
\newblock \bibinfo{journal}{Advances in neural information processing systems} \bibinfo{volume}{29} (\bibinfo{year}{2016}).
\bibitem[{Bartl et~al.(2020)Bartl, Nissim, and Gatt}]{bartl2020unmasking}
\bibinfo{author}{M.~Bartl}, \bibinfo{author}{M.~Nissim}, \bibinfo{author}{A.~Gatt},
\newblock \bibinfo{title}{Unmasking contextual stereotypes: Measuring and mitigating bert’s gender bias},
\newblock in: \bibinfo{booktitle}{Proceedings of the Second Workshop on Gender Bias in Natural Language Processing}, \bibinfo{year}{2020}, pp. \bibinfo{pages}{1--16}.
\bibitem[{Piazzolla et~al.(2023)Piazzolla, Savoldi, and Bentivogli}]{piazzolla2023good}
\bibinfo{author}{S.~A. Piazzolla}, \bibinfo{author}{B.~Savoldi}, \bibinfo{author}{L.~Bentivogli},
\newblock \bibinfo{title}{Good, but not always fair: An evaluation of gender bias for three commercial machine translation systems},
\newblock \bibinfo{journal}{HERMES}  (\bibinfo{year}{2023}) \bibinfo{pages}{209--225}.
\bibitem[{Zhao et~al.(2020)Zhao, Mukherjee, Chang, Hassan~Awadallah et~al.}]{zhao2020gender}
\bibinfo{author}{J.~Zhao}, \bibinfo{author}{S.~Mukherjee}, \bibinfo{author}{K.-W. Chang}, \bibinfo{author}{A.~Hassan~Awadallah}, et~al.,
\newblock \bibinfo{title}{Gender bias in multilingual embeddings and cross-lingual transfer},
\newblock in: \bibinfo{booktitle}{Proceedings of the 58th Annual Meeting of the Association for Computational Linguistics}, \bibinfo{year}{2020}.
\bibitem[{Zhao et~al.(2024)Zhao, Ding, Jia, Wang, and Qian}]{zhao2024gender}
\bibinfo{author}{J.~Zhao}, \bibinfo{author}{Y.~Ding}, \bibinfo{author}{C.~Jia}, \bibinfo{author}{Y.~Wang}, \bibinfo{author}{Z.~Qian},
\newblock \bibinfo{title}{Gender bias in large language models across multiple languages},
\newblock \bibinfo{journal}{arXiv preprint arXiv:2403.00277}  (\bibinfo{year}{2024}).
\bibitem[{Stanovsky et~al.(2019)Stanovsky, Smith, and Zettlemoyer}]{stanovsky2019evaluating}
\bibinfo{author}{G.~Stanovsky}, \bibinfo{author}{N.~A. Smith}, \bibinfo{author}{L.~Zettlemoyer},
\newblock \bibinfo{title}{Evaluating gender bias in machine translation},
\newblock in: \bibinfo{booktitle}{Proceedings of the 57th Annual Meeting of the Association for Computational Linguistics}, \bibinfo{year}{2019}, pp. \bibinfo{pages}{1679--1684}.
\bibitem[{Prates et~al.(2020)Prates, Avelar, and Lamb}]{prates2020assessing}
\bibinfo{author}{M.~O. Prates}, \bibinfo{author}{P.~H. Avelar}, \bibinfo{author}{L.~C. Lamb},
\newblock \bibinfo{title}{Assessing gender bias in machine translation: a case study with google translate},
\newblock \bibinfo{journal}{Neural Computing and Applications} \bibinfo{volume}{32} (\bibinfo{year}{2020}) \bibinfo{pages}{6363--6381}.
\bibitem[{Savoldi et~al.(2021)Savoldi, Gaido, Bentivogli, Negri, and Turchi}]{savoldi2021gender}
\bibinfo{author}{B.~Savoldi}, \bibinfo{author}{M.~Gaido}, \bibinfo{author}{L.~Bentivogli}, \bibinfo{author}{M.~Negri}, \bibinfo{author}{M.~Turchi},
\newblock \bibinfo{title}{Gender bias in machine translation},
\newblock \bibinfo{journal}{Transactions of the Association for Computational Linguistics} \bibinfo{volume}{9} (\bibinfo{year}{2021}) \bibinfo{pages}{845--874}. \DOIprefix\doi{10.1162/tacl_a_00401}.
\bibitem[{Cho et~al.(2019)Cho, Kim, Kim, and Kim}]{cho2019measuring}
\bibinfo{author}{W.~I. Cho}, \bibinfo{author}{J.~W. Kim}, \bibinfo{author}{S.~M. Kim}, \bibinfo{author}{N.~S. Kim},
\newblock \bibinfo{title}{On measuring gender bias in translation of gender-neutral pronouns},
\newblock in: \bibinfo{booktitle}{Proceedings of the First Workshop on Gender Bias in Natural Language Processing}, \bibinfo{year}{2019}, pp. \bibinfo{pages}{173--181}.
\bibitem[{Webber et~al.(2010)Webber, Moffat, and Zobel}]{webber2010similarity}
\bibinfo{author}{W.~Webber}, \bibinfo{author}{A.~Moffat}, \bibinfo{author}{J.~Zobel},
\newblock \bibinfo{title}{A similarity measure for indefinite rankings},
\newblock \bibinfo{journal}{ACM Transactions on Information Systems (TOIS)} \bibinfo{volume}{28} (\bibinfo{year}{2010}) \bibinfo{pages}{1--38}.
\bibitem[{Yang et~al.(2020)Yang, Cer, Ahmad, Guo, Law, Constant, Abrego, Yuan, Tar, Sung et~al.}]{yang2020multilingual}
\bibinfo{author}{Y.~Yang}, \bibinfo{author}{D.~Cer}, \bibinfo{author}{A.~Ahmad}, \bibinfo{author}{M.~Guo}, \bibinfo{author}{J.~Law}, \bibinfo{author}{N.~Constant}, \bibinfo{author}{G.~H. Abrego}, \bibinfo{author}{S.~Yuan}, \bibinfo{author}{C.~Tar}, \bibinfo{author}{Y.-H. Sung}, et~al.,
\newblock \bibinfo{title}{Multilingual universal sentence encoder for semantic retrieval},
\newblock in: \bibinfo{booktitle}{Proceedings of the 58th Annual Meeting of the Association for Computational Linguistics: System Demonstrations}, \bibinfo{year}{2020}, pp. \bibinfo{pages}{87--94}.
\bibitem[{Wang et~al.(2024)Wang, Yang, Huang, Yang, Majumder, and Wei}]{wang2024multilingual}
\bibinfo{author}{L.~Wang}, \bibinfo{author}{N.~Yang}, \bibinfo{author}{X.~Huang}, \bibinfo{author}{L.~Yang}, \bibinfo{author}{R.~Majumder}, \bibinfo{author}{F.~Wei},
\newblock \bibinfo{title}{Multilingual e5 text embeddings: A technical report},
\newblock \bibinfo{journal}{arXiv preprint arXiv:2402.05672}  (\bibinfo{year}{2024}).
\bibitem[{Reimers and Gurevych(2019)}]{reimers-2019-sentence-bert}
\bibinfo{author}{N.~Reimers}, \bibinfo{author}{I.~Gurevych},
\newblock \bibinfo{title}{Sentence-bert: Sentence embeddings using siamese bert-networks},
\newblock in: \bibinfo{booktitle}{Proceedings of the 2019 Conference on Empirical Methods in Natural Language Processing}, \bibinfo{publisher}{Association for Computational Linguistics}, \bibinfo{year}{2019}. \URLprefix \url{http://arxiv.org/abs/1908.10084}.

\end{thebibliography}

\appendix

\section{Additional Results}
\label{sec:appendix}

Table~\ref{tab:appendix_results_rbo} shows the RBO results when comparing the ranks using masculine versus feminine input queries over the same corpora, in each of our covered languages. The results in this table are comparable to those in Table~\ref{tab:results} in Section~\ref{sec:results}, but in the current table we use the corpora of feminine job titles instead of masculine ones.

\begin{table}[h]
\centering
\caption{RBO results of zero-shot performance of selected models, comparing feminine queries versus masculine queries over feminine corpora. Higher scores, in bold, mean less gender impact, and lower scores, underlined, a higher impact of changing the queries' gender.}  
\label{tab:appendix_results_rbo}
\begin{tabular}{|l|c|c|c|c|}
\hline
\textbf{Model}   & \textbf{DE}                 & \textbf{ES}                          & \textbf{FR}                          & \textbf{PT}                          \\ \hline
\textbf{m-e5}    & \underline{0.8316}             & 0.8755                               & \underline{0.8683}                      & \underline{0.8720}                      \\ \hline
\textbf{m-e5-l}  & 0.8635                      & \underline{0.8678}                      & 0.8795                               & 0.8747                               \\ \hline
\textbf{m-use}   & 0.9084                      & 0.9038                               & 0.9234                               & 0.9006                               \\ \hline
\textbf{m-use-l} & \textbf{0.9199}             & 0.9092                               & 0.9420                               & 0.9274                               \\ \hline
\textbf{minilm}  & \multicolumn{1}{l|}{0.9008} & \multicolumn{1}{l|}{\textbf{0.9433}} & \multicolumn{1}{l|}{\textbf{0.9453}} & \multicolumn{1}{l|}{\textbf{0.9511}} \\ \hline

\end{tabular}
\end{table}

The results show patterns similar to those obtained on the masculine corpora, but with consistently lower results, indicating a more disruptive effect of using feminine inputs. 
Again, 
minilm obtains the highest RBO scores except for German, for which it shows a greater sensibility than for the other languages, but every model tested for this task exhibits gender bias.


As we believe that measuring gender bias is an additional metric to model performance, we include the Mean Average Precision (MAP) scores obtained when evaluating the selected models over the masculine and feminine corpora in Tables~\ref{tab:map_m} and \ref{tab:map_f} respectively, using the feminine and masculine queries for each covered language.

\begin{table*}
\centering
\caption{MAP scores over the masculine corpora.}
\label{tab:map_m}
\begin{tabular}{|l|cc|cc|cc|cc|}
\hline
\textbf{}                & \multicolumn{2}{c|}{\textbf{DE}}                       & \multicolumn{2}{c|}{\textbf{ES}}                       & \multicolumn{2}{c|}{\textbf{FR}}                       & \multicolumn{2}{c|}{\textbf{PT}}                       \\ \cline{2-9} 
\textbf{Model / Queries} & \multicolumn{1}{c|}{\textbf{F}}      & \textbf{M}      & \multicolumn{1}{c|}{\textbf{F}}      & \textbf{M}      & \multicolumn{1}{c|}{\textbf{F}}      & \textbf{M}      & \multicolumn{1}{c|}{\textbf{F}}      & \textbf{M}      \\ \hline
\textbf{m-e5}            & \multicolumn{1}{c|}{0.3178}          & 0.3237          & \multicolumn{1}{c|}{0.4620}          & 0.4845& \multicolumn{1}{c|}{0.3422}          & 0.3547          & \multicolumn{1}{c|}{0.3470}          & 0.3489          \\ \hline
\textbf{m-e5-l}          & \multicolumn{1}{c|}{\textbf{0.3463}} & \textbf{0.3523} & \multicolumn{1}{c|}{\textbf{0.4644}} & \textbf{0.4862}& \multicolumn{1}{c|}{\textbf{0.3709}} & \textbf{0.3803} & \multicolumn{1}{c|}{\textbf{0.3597}} & \textbf{0.3603} \\ \hline
\textbf{m-use}           & \multicolumn{1}{c|}{0.3102}          & 0.3123          & \multicolumn{1}{c|}{0.4490}          & 0.4779& \multicolumn{1}{c|}{0.3298}          & 0.3490          & \multicolumn{1}{c|}{0.3495}          & 0.3534          \\ \hline
\textbf{m-use-l}         & \multicolumn{1}{c|}{0.3190}          & 0.3233          & \multicolumn{1}{c|}{0.4307}          & 0.4616& \multicolumn{1}{c|}{\underline{0.3210}} & \underline{0.3350} & \multicolumn{1}{c|}{0.3424}          & 0.3400          \\ \hline
\textbf{minilm}          & \multicolumn{1}{c|}{\underline{0.2591}} & \underline{0.2635} & \multicolumn{1}{c|}{\underline{0.4253}} & \underline{0.4456}& \multicolumn{1}{c|}{0.3328}          & 0.3500          & \multicolumn{1}{c|}{\underline{0.3142}} & \underline{0.3149} \\ \hline   

\end{tabular}
\end{table*}

\begin{table*}
\centering
\caption{MAP scores over the feminine corpora.}
\label{tab:map_f}
\begin{tabular}{|l|cc|cc|cc|cc|}
\hline
\textbf{}                     & \multicolumn{2}{c|}{\textbf{DE}}                       & \multicolumn{2}{c|}{\textbf{ES}}                       & \multicolumn{2}{c|}{\textbf{FR}}                       & \multicolumn{2}{c|}{\textbf{PT}}                       \\ \cline{2-9} 
\textbf{Model    /   Queries} & \multicolumn{1}{c|}{\textbf{F}}      & \textbf{M}      & \multicolumn{1}{c|}{\textbf{F}}      & \textbf{M}      & \multicolumn{1}{c|}{\textbf{F}}      & \textbf{M}      & \multicolumn{1}{c|}{\textbf{F}}      & \textbf{M}      \\ \hline

\textbf{m-e5}            & \multicolumn{1}{c|}{0.2960}          & 0.3260          & \multicolumn{1}{c|}{\textbf{0.4727}} & 0.4618          & \multicolumn{1}{c|}{0.3320}          & 0.3540          & \multicolumn{1}{c|}{0.3378}          & 0.3557          \\ \hline
\textbf{m-e5-l}          & \multicolumn{1}{c|}{\textbf{0.3326}} & \textbf{0.3606} & \multicolumn{1}{c|}{0.4663}          & \textbf{0.4696} & \multicolumn{1}{c|}{\textbf{0.3587}} & \textbf{0.3807} & \multicolumn{1}{c|}{0.3453}          & \textbf{0.3628} \\ \hline
\textbf{m-use}           & \multicolumn{1}{c|}{0.2989}          & 0.3045          & \multicolumn{1}{c|}{0.4704}          & 0.4394          & \multicolumn{1}{c|}{0.3346}          & 0.3479          & \multicolumn{1}{c|}{\textbf{0.3459}} & 0.3494          \\ \hline
\textbf{m-use-l}         & \multicolumn{1}{c|}{0.3132}          & 0.3176          & \multicolumn{1}{c|}{0.4543} & 0.4178 & \multicolumn{1}{c|}{\underline{0.3200}} & \underline{0.3277} & \multicolumn{1}{c|}{0.3388}          & 0.3333          \\ \hline
\textbf{minilm}          & \multicolumn{1}{c|}{\underline{0.2550}} & \underline{0.2516} & \multicolumn{1}{c|}{\underline{0.4448}}          & \underline{0.4153}          & \multicolumn{1}{c|}{0.3305}          & 0.3512          & \multicolumn{1}{c|}{\underline{0.3243}} & \underline{0.3253} \\ \hline

\end{tabular}
\end{table*}

As shown by the RBO scores, the detailed MAP results also provide evidence of gender bias, since we get different scores when the data differ only by gender. In general, the difference between masculine and feminine tends to be larger for German, which is also the language on which the models perform the worst, but performance gaps can be found for the other languages as well. 

\section{Sample Rankings}
\label{sec:appendix_b}

Table~\ref{tab:rank_lawyer_es} shows the top 20 ranked job titles for the same query (`lawyer') in its masculine versus feminine form in Spanish (`abogado', `abogada'), when using the m-e5 model to match job titles from the same corpus (in this case, including masculine job titles). We can see that the rankings are not the same when the only difference is the marked gender in the input query. This denotes gender bias in the job title matching task, as we understand that a fair system would give similar opportunities to candidates with the same job title, whether it is expressed in its feminine or masculine form.

\begin{table}[]
\label{tab:rank_lawyer_es}
\caption{Top 20 ranking results for the same query `lawyer' in Spanish in feminine versus in masculine, over the same corpus of masculine job titles.}
\begin{tabular}{ccc}
\hline
\multicolumn{1}{|c|}{\textbf{Query}} & \multicolumn{1}{c|}{\textbf{abogada (f)}}              & \multicolumn{1}{c|}{\textbf{abogado (m)}}              \\ \hline
\multicolumn{1}{|c|}{1}              & \multicolumn{1}{c|}{abogado adjunto}                & \multicolumn{1}{c|}{abogado}                        \\ \hline
\multicolumn{1}{|c|}{2}              & \multicolumn{1}{c|}{abogado}                        & \multicolumn{1}{c|}{abogado de tribunales}          \\ \hline
\multicolumn{1}{|c|}{3}              & \multicolumn{1}{c|}{asociado}                       & \multicolumn{1}{c|}{abogado adjunto}                \\ \hline
\multicolumn{1}{|c|}{4}              & \multicolumn{1}{c|}{agente judicial}                & \multicolumn{1}{c|}{abogado litigante}              \\ \hline
\multicolumn{1}{|c|}{5}              & \multicolumn{1}{c|}{asociado jurídico}              & \multicolumn{1}{c|}{abogado de empresa}             \\ \hline
\multicolumn{1}{|c|}{6}              & \multicolumn{1}{c|}{asistente judicial}             & \multicolumn{1}{c|}{abogado de reclamaciones}       \\ \hline
\multicolumn{1}{|c|}{7}              & \multicolumn{1}{c|}{abogado de tribunales}          & \multicolumn{1}{c|}{abogado externo}                \\ \hline
\multicolumn{1}{|c|}{8}              & \multicolumn{1}{c|}{abogado litigante}              & \multicolumn{1}{c|}{abogado especial}               \\ \hline
\multicolumn{1}{|c|}{9}              & \multicolumn{1}{c|}{defensor público adjunto}       & \multicolumn{1}{c|}{abogado de derecho de familia}  \\ \hline
\multicolumn{1}{|c|}{10}             & \multicolumn{1}{c|}{asesor jurídico}                & \multicolumn{1}{c|}{ayudante del abogado de oficio} \\ \hline
\multicolumn{1}{|c|}{11}             & \multicolumn{1}{c|}{abogado gerente}                & \multicolumn{1}{c|}{abogado senior}                 \\ \hline
\multicolumn{1}{|c|}{12}             & \multicolumn{1}{c|}{urbanista}                      & \multicolumn{1}{c|}{abogado gerente}                \\ \hline
\multicolumn{1}{|c|}{13}             & \multicolumn{1}{c|}{ayudante del abogado de oficio} & \multicolumn{1}{c|}{abogado de contratos}           \\ \hline
\multicolumn{1}{|c|}{14}             & \multicolumn{1}{c|}{abogado de derecho de familia}  & \multicolumn{1}{c|}{abogado investigador}           \\ \hline
\multicolumn{1}{|c|}{15}             & \multicolumn{1}{c|}{abogado senior}                 & \multicolumn{1}{c|}{abogado penalista}              \\ \hline
\multicolumn{1}{|c|}{16}             & \multicolumn{1}{c|}{juez de derecho administrativo} & \multicolumn{1}{c|}{abogado laboralista}            \\ \hline
\multicolumn{1}{|c|}{17}             & \multicolumn{1}{c|}{abogado corporativo adjunto}    & \multicolumn{1}{c|}{abogado inmobiliario}           \\ \hline
\multicolumn{1}{|c|}{18}             & \multicolumn{1}{c|}{abogado laboralista}            & \multicolumn{1}{c|}{abogado fiscal}                 \\ \hline
\multicolumn{1}{|c|}{19}             & \multicolumn{1}{c|}{abogado externo}                & \multicolumn{1}{c|}{abogado internacional}          \\ \hline
\multicolumn{1}{|c|}{20}             & \multicolumn{1}{c|}{becario jurídico}               & \multicolumn{1}{c|}{abogado corporativo}            \\ \hline                               
\end{tabular}
\end{table}

\end{document}